\documentclass[sigconf,nonacm]{aamas} 

\usepackage{balance} 

\usepackage{multirow}
\usepackage{graphicx}

\usepackage{amsmath}
\usepackage{amsthm}
\theoremstyle{plain}
\usepackage{subcaption}
\usepackage{algorithm}
\usepackage{algorithmic}
\usepackage{threeparttable}
\newcommand{\setlabel}[1]{\edef\@currentlabel{#1}\label}

\theoremstyle{acmplain}

\newtheorem{re}{Theorem}
\newtheorem{pos}{Theorem}
\newtheorem{defi}{Theorem}
\newtheorem{remark}[re]{Remark}
\newtheorem{proposition}[pos]{Proposition}
\newtheorem{definition}[defi]{Definition}

\usepackage{import}
\mathchardef\mhyphen="2D

\def\cA{\mathcal{A}}
\def\cB{\mathcal{B}}

\DeclareMathOperator{\SW}{\texttt{SW}}
\DeclareMathOperator{\SSW}{\texttt{SSW}}
\usepackage{pifont}
\newcommand{\cmark}{\ding{51}} 
\newcommand{\xmark}{\ding{55}} 

\newif\ifshowappendix
\showappendixtrue   
\ifshowappendix
\else
    \usepackage{xr-hyper}
    \usepackage{hyperref} 
\fi

\makeatletter
\gdef\@copyrightpermission{
  \begin{minipage}{0.2\columnwidth}
   \href{https://creativecommons.org/licenses/by/4.0/}{\includegraphics[width=0.90\textwidth]{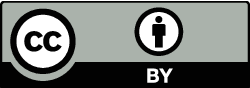}}
  \end{minipage}\hfill
  \begin{minipage}{0.8\columnwidth}
   \href{https://creativecommons.org/licenses/by/4.0/}{This work is licensed under a Creative Commons Attribution International 4.0 License.}
  \end{minipage}
  \vspace{5pt}
}
\makeatother

\setcopyright{ifaamas}
\acmConference[AAMAS '25]{Proc.\@ of the 24th International Conference
on Autonomous Agents and Multiagent Systems (AAMAS 2025)}{May 19 -- 23, 2025}
{Detroit, Michigan, USA}{Y.~Vorobeychik, S.~Das, A.~Nowé  (eds.)}
\copyrightyear{2025}
\acmYear{2025}
\acmDOI{}
\acmPrice{}
\acmISBN{}

\acmSubmissionID{<<1303>>}

\title{\texttt{DUPRE}: Data Utility Prediction for Efficient Data Valuation}

\author{Kieu Thao Nguyen Pham}
\authornote{Equal contribution.}
\affiliation{
  \institution{National University of Singapore}
  \country{Singapore}}
\email{nguyen.pkt@u.nus.edu}

\author{Rachael Hwee Ling Sim}
\authornotemark[1]
\affiliation{
  \institution{National University of Singapore}
  \country{Singapore}}
\email{rachael.sim@u.nus.edu}

\author{Quoc Phong Nguyen}
\affiliation{
  \institution{A2I2, Deakin University}
  \country{Australia}}
\email{ qphongmp@gmail.com}

\author{See Kiong Ng}
\affiliation{
  \institution{National University of Singapore}
  \country{Singapore}}
\email{seekiong@nus.edu.sg}

\author{Bryan Kian Hsiang Low}
\affiliation{
  \institution{National University of Singapore}
  \country{Singapore}}
\email{lowkh@comp.nus.edu.sg}

\begin{abstract}
Data valuation is increasingly used in machine learning (ML) to decide the fair compensation for data owners and identify valuable or harmful data for improving ML models. Cooperative game theory-based data valuation, such as Data Shapley, requires evaluating the data utility (e.g., validation accuracy) and retraining the ML model for multiple data subsets. While most existing works on efficient estimation of the Shapley values have focused on reducing the number of subsets to evaluate, our framework, \texttt{DUPRE}, takes an alternative yet complementary approach that reduces the cost per subset evaluation by predicting data utilities instead of evaluating them by model retraining. Specifically, given the evaluated data utilities of some data subsets, \texttt{DUPRE} fits a \emph{Gaussian process} (GP) regression model to predict the utility of every other data subset. Our key contribution lies in the design of our GP kernel based on the sliced Wasserstein distance between empirical data distributions. In particular, we show that the kernel is valid and positive semi-definite, encodes prior knowledge of similarities between different data subsets, and can be efficiently computed. We empirically verify that \texttt{DUPRE} introduces low prediction error and speeds up data valuation for various ML models, datasets, and utility functions.

\end{abstract}

\newcommand{\BibTeX}{\rm B\kern-.05em{\sc i\kern-.025em b}\kern-.08em\TeX}

\begin{document}


\pagestyle{fancy}
\fancyhead{}


\maketitle 

\section{Introduction} \label{sec:intro}
In recent years, there has been growing interest in \emph{data valuation} and understanding how much data is worth in \emph{machine learning}~(ML). Data valuation can be used to determine the fair compensation that data owners deserve for sharing their data \cite{jia2019towards,sim2020cml} and to identify valuable datasets to explain and improve the performance of their models~\cite{ghorbani2019data}.
A common category of data valuation methods that values a data point/set (relative to the data contributed by others) is \emph{cooperative game theory} (CGT) based valuation \cite{sim2022data}. 
Suppose the ML model is trained on data from a set $N$ of $n$ data owners (owners).
Data Shapley \cite{ghorbani2019data}, a popular CGT-based valuation technique, values an owner $i$ by its Shapley value
\begin{equation}\label{eq:shapley}
    \phi_u(i) \triangleq \sum_{C \subseteq N \setminus i} \frac{1}{n} {\binom{n-1}{|C|}}^{-1} \left[ u(C \cup \{i\}) - u(C) \right].
\end{equation}

The data utility function $u$ maps any coalition (i.e., set) $C \subseteq N$ of owners to the ML performance achievable by their data. A concrete example of $u$ is the validation accuracy on a trained neural network.
Other CGT-based valuations include the least-core solution \cite{yan2021core} and semivalues \cite{Kwon2021betashapley,wang2023data}, such as the Banzhaf value, which is similar to the Shapley value but uses a different set of weights $(\omega_c)_{c=0}^{n-1}$ such that each $\omega_c \geq 0$ and $\sum_{c=0}^{n-1} \omega_{c}{\binom{n-1}{c}}=1$ in the following expression
\begin{equation}\label{eq:semi}
\varphi_u(i) \triangleq \sum_{C \subseteq N \setminus i}\omega_{|C|} \left[ u(C \cup \{i\}) - u(C) \right].
\end{equation}

While CGT-based valuations satisfy desirable axioms, they all require an exponential number of evaluations of~$u$.
This high complexity motivates existing works to study more efficient methods for estimating the Shapley value and other semivalues by~\textbf{(i)} reducing the number of evaluations or~\textbf{(ii)} reducing the cost per evaluation. Each data utility evaluation may involve expensive model retraining from scratch to evaluate its predictive performance. Our work addresses~\textbf{(ii)} and is complementary to methods that address~\textbf{(i)}.

Most existing works focus on \textbf{(i)} and propose Monte Carlo methods, such as permutation sampling, group testing \cite{jia2019towards}, and reusing samples (data utility evaluations) to compute the semivalues for multiple owners efficiently \cite{LiYu24,kolpaczki2024without}. 
To address \textbf{(ii)}, some works propose heuristics to avoid retraining the ML model from scratch. 
For example, TMC-Shapley~\cite{ghorbani2019data} simply approximates $\Delta_u(i|C) \triangleq u(C \cup i) - u(C)$ with $0$ when $u(C)$ is sufficiently close to $u(N)$.
Gradient Shapley~\cite{ghorbani2019data} considers training the ML model (e.g., deep neural network) over a single training epoch so multiple utilities (e.g., $u(C)$, $u(C \cup i)$ then $u(C \cup \{i,j\})$) can be incrementally computed. ~\citet{jia2019towards} suggest using the influence function heuristic to approximate $\Delta_u(i|C)$ when owner $i$ owns a single data point. It also describes how to compute the Shapley value of all points exactly in log-linear time for their $k$-nearest neighbor utility function.
However, these methods cannot be applied for \emph{all} models (e.g., neural networks trained over \emph{multiple} epochs), utility functions (e.g., when $u(C) \ll u(N)$) and datasets (e.g., the influence approximation may be inaccurate when each owner owns a larger dataset instead \cite{koh2019accuracy}).
These limitations raise an important question: For faster data valuation, is there a \emph{general} method to \emph{reduce the cost per data utility evaluations} that works for \emph{all} models, utility functions, and datasets?

\citet{wang2021predict} have proposed predicting data utility for any input dataset and using a hybrid of actual utility evaluations and predicted evaluations during data valuation. They briefly analyzed how this hybrid slightly worsens the CGT-based valuation approximation error. However, \citet{wang2021predict} did not optimize the predictor and simply trained a neural network that takes in an indicator vector for each coalition $C$ (i.e., an $n$-dimensional vector where each entry is $1$ if the corresponding owner is present in $C$ and $0$ if absent).

Our work seeks to optimize the predictor and addresses the following questions: (1) Can the predictor further exploit the similarity between data of different owners? For example, if owners~$i$ and~$j$ have highly similar data, can the predictor leverage the prior knowledge that $u(C \cup i) \approx u(C \cup j)$ (instead of learning the relationship from more data utility evaluations)? (2) Additionally, can the predictor quantify the uncertainty in its prediction?

\emph{Gaussian process regression} (GPR) seems well-suited to this task. We can (1) incorporate prior knowledge by specifying an appropriate kernel over datasets and (2) quantify the additional uncertainty in the estimated semivalue (due to the predictor) using the GPR posterior covariance.
However, adapting GPR presents challenges: The kernel must be \emph{positive semi-definite} (PSD), to ensure a nondegenerate GP posterior and offer computational savings as compared to directly evaluating the utility from model retraining.
We propose measuring the similarity between empirical data distributions with a \emph{sliced Wasserstein distance} ($\SW$) kernel, as \citet{meunier2022slicedw} have proven that the kernel is PSD. 
Additionally, our method is computationally efficient: after precomputing the sorted projections once, the $\SW$ distance used in each kernel entry can be computed in linear time w.r.t.~the total dataset size. 
Our method, \texttt{DUPRE}, fits a GP model based on some actual data utility evaluations. Subsequently, \texttt{DUPRE} predicts the utility of unevaluated coalitions and can be used to identify those with high uncertainty for further actual evaluations. \texttt{DUPRE} complements any exact CGT-based valuation and approximation techniques proposed to address \textbf{(i)}.

In summary, we make the following key contributions.
In Section~\ref{sec:pf}, we formulate the problem of predicting the utility of coalitions using GPR and propose a suitable valid kernel to measure the similarity between datasets for regression and classification problems. In Section~\ref{sec:semi}, we describe how to estimate the semivalues based on the GP model.\footnote{The supplementary materials can be found at \url{https://kakaeriol.github.io/dupre/}.} In Section~\ref{sec:exp}, we empirically verify that \texttt{DUPRE} introduces a low prediction error and speeds up data valuation for various few models, datasets, and utility functions. 

\section{Problem formulation} 
\label{sec:pf}
We consider $n$ data owners $N = \{1, \dots, n\}$. Data owner $i$ has a dataset $D_i \triangleq (X_i, y_i)$ where $X_i$ is the input matrix and $y_i$ is the target outputs. Each owner $i$ shares their dataset $D_i$ with the mediator, who train ML models on data from multiple owners to assign a fair data value $\phi_i$ to each data owner $i$.
Let $C \subseteq N $ denote a \emph{coalition} of data owners with the aggregated dataset~$D_C \triangleq (X_C, y_C)$, where~$X_C \triangleq \cup_{i \in C} X_i$ and~$y_C \triangleq \cup_{i \in C} y_i$.\footnote{For notation convenience, we say $(x,y) \in D_C$ if there exists an index $j$ such that the $j$ element of $X_C$ and $y_C$ are $x$ and $y$, respectively.}

\begin{figure}[!ht]
    \centering
    \includegraphics[width=\columnwidth]{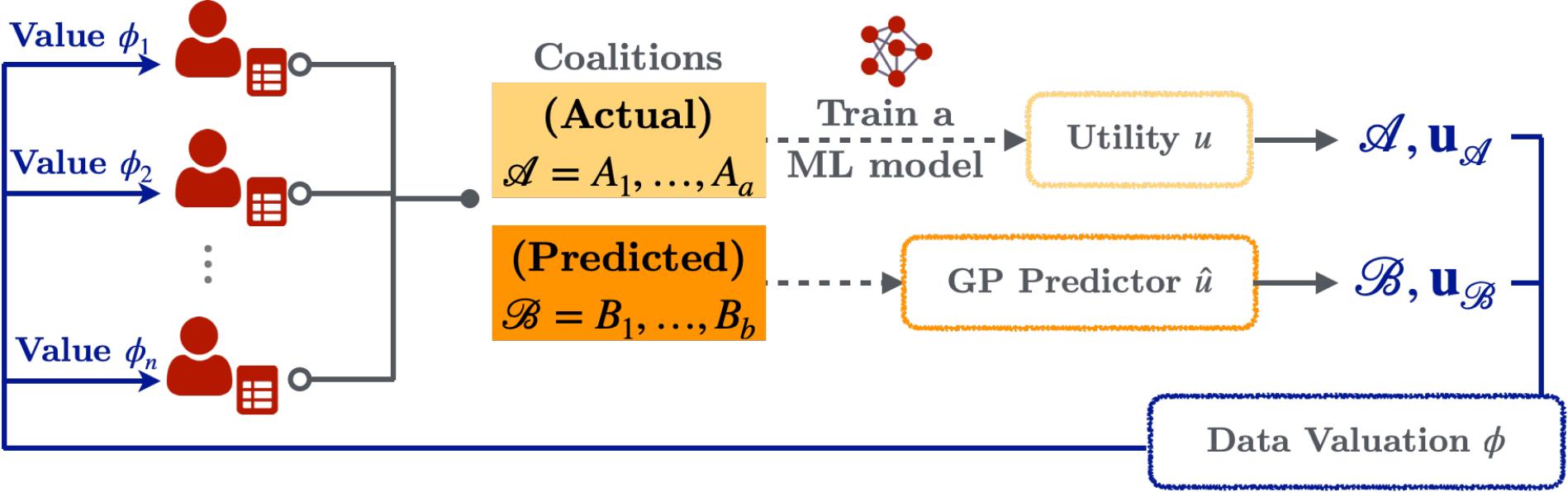}
    \caption{We partition all (sampled) coalitions into two families of sets. The utilities of coalitions in  $\cA$ are actually evaluated by training the ML model and measuring the utility, e.g., validation accuracy. In contrast, the utilities of coalitions in $\cB$ are predicted by the Gaussian Process (GP) model.
    }
    \label{fig:framework}
    \Description{DUPRE framework}
\end{figure}
The data utility function $u$ maps any coalition $C$ to the performance (e.g., negated mean squared error) of the ML model trained on their data, $u(C)$. The function $u$ may be expensive to evaluate for complex models such as deep neural networks.

For any $j$, let coalitions $A_j$ and $B_j$ be subsets of $N$.
Given $a$ actual utility evaluations of coalitions in $\cA \triangleq (A_j)_{j=1}^a$ (i.e., $(u(A))_{A \in \cA}$), our goal is to learn a predictor $\hat{u}$ that predicts $b$ data utilities of coalitions in $\cB \triangleq (B_j)_{j=1}^b$.
Subsequently, we use both the actual utility evaluations at $\cA$ and predicted utility evaluations at $\cB$ to exactly compute or approximate the CGT-based data valuation, as shown in Figure~\ref{fig:framework}.

For each coalition $C$, we model the data utility $u(C) = \upsilon(C) + \epsilon_C$ where $\epsilon_C$ is sampled from a Gaussian distribution with zero mean and variance $\sigma^2$. We specify the underlying generating function $\upsilon$ as a \emph{Gaussian process} \cite{williams_gaussian_1996} with a covariance kernel $k$ (see Appendix~\ref{appendix:gp_background}). Given the observed utilities $\mathbf{u}_\cA$ for coalitions $\cA$, the posterior belief of the utilities $\hat{\mathbf{u}}_{\cB|\cA}$ for coalitions $\cB$ follows a Gaussian distribution
\begin{align*}
  \mathbf{\hat{u}}_{\cB|\cA} & \sim \mathcal{N}
  \left(\mathbb{E}[\mathbf{\hat{u}}_{\cB|\cA}], \mathbb{V}[\mathbf{\hat{u}}_{\cB|\cA}]   \right) \\
  \mathbb{E}[\mathbf{\hat{u}}_{\cB|\cA}] &=
  \mathbf{K}_{\cB,\cA} [\mathbf{K}_{\cA,\cA} + \sigma^2 \mathbf{I}]^{-1} \mathbf{u}_\cA \\ 
  \mathbb{V}[\mathbf{\hat{u}}_{\cB|\cA}] &= 
  \mathbf{K}_{\cB,\cB} - \mathbf{K}_{\cB,\cA}  [\mathbf{K}_{\cA,\cA} + \sigma^2 \mathbf{I}]^{-1} \mathbf{K}_{\cA, \cB}\ .
\end{align*}

Here, $\mathbf{K}_{\cB,\cA}$ is a matrix whose $j,i$ entry is $k(B_j, A_i)$, the similarity between the aggregated datasets of $D_{B_j}$ and that of $D_{A_i}$. Notice that there are $O(a^2 + ab + b^2)$ unique kernel entries in  $\mathbf{K}_{\cB,\cA}$ and  $\mathbf{K}_{\cA,\cA}$.
The inverse of the kernel matrix only needs to be computed once in $O(a^3)$ time. Subsequently, each utility prediction only involves matrix multiplication.

Our next challenge is to decide the kernel function $k$ over datasets or data distributions such that the kernel is \textbf{(I)} valid (see Appendix~\ref{appendix:pos} for properties a valid kernel must satisfy such as positive semi-definite, PSD) and \textbf{(II)} computationally efficient. The former results in a valid GP while the latter ensures that the method is useful in practice. 

\subsection{Choice of Kernel}
For simplicity, we start by ignoring the target outputs and considering only the input matrix, i.e., $D_i = X_i$ for every $i$.
How do we measure the similarity between the aggregated dataset $D_A$ (from owners in $A$) and $D_B$? Equivalently, let $\delta$ be the Dirac delta distribution, how do we measure the distance between the empirical data distributions $\mathbb{P}(D_A) =\frac{1}{|D_A|}\sum_{x \in D_A}{\delta (x)}$ and $\mathbb{P}(D_B)$?
We quantify the distance between the empirical data distributions using optimal transport distances (see Appendix~\ref{appendix:otdd_background}) rather than f-divergences. Optimal transport distances, such as the Wasserstein distance, exhibit desirable mathematical properties including symmetry and the triangle inequality, which are essential for defining valid kernels and comparing distributions even when their supports are disjoint. Optimal transport distances measure the minimal total cost required to transform one distribution into another. In particular, the Wasserstein distance captures the intrinsic geometry of the space of distributions \cite{meunier2022slicedw} and admits an intuitive interpretation: it is the minimum total cost of transporting mass from the distribution $\mathbb{P}(D_A)$ to $\mathbb{P}(D_B)$.

However, the kernel based on the squared Wasserstein distance, i.e., $k(A, B) \propto e^{-W_2^2(\mathbb{P}(D_A), \mathbb{P}(D_B))}$ may not be PSD when the data dimension exceeds $1$ \cite{Ginsbourger2018GaussianPW,meunier2022slicedw}, thus violating \textbf{(I)}.
Moreover, as computing the Wasserstein distance involves an optimization problem, the most efficient method \cite{dvurechensky2018compot} still takes $\Tilde{O}(\max(|D_A|,|D_B|)^2)$ time.\footnote{$\Tilde{O}$ hides polylogarithmic factors.} This computation burden becomes expensive when repeated $O(a^2 + ab + b^2)$ times for each kernel entry, violating \textbf{(II)}.
Thus, we must use alternatives to the Wasserstein distance, such as the \emph{sliced Wasserstein distance} (SW) \cite{meunier2022slicedw} that provably satisfies \textbf{(I)}.

\begin{proposition}[\cite{meunier2022slicedw}]\label{prop:sw}
    The exponential kernels $k(A, B)$ based on the $\SW$ distance, including 
    $\exp\left(-\gamma \cdot \SW_2^{2\rho}\left(\mathbb{P}(D_A), \mathbb{P}(D_B)\right)\right)$ and \linebreak
    $\exp\left(-\gamma \cdot \SW_1^{\rho}\left(\mathbb{P}(D_A), \mathbb{P}(D_B)\right)\right)$ 
    are positive semi-definite (PSD) and valid for $\gamma > 0$ and $\rho \in [0,1]$.
\end{proposition}

To address \textbf{(II)}, the SW distance can be efficiently approximated using Monte Carlo sampling with $L$ projections. After each projection, the Wasserstein distance between one-dimensional distributions can be computed analytically.
We additionally observe that only line~10 in Algorithm~\ref{alg:SW}, which merges two sorted projection lists, is unique to the coalition pair $(A, B)$. Thus, only this step, which takes $O(L \cdot (|D_A|+|D_B|))$ time, is repeated for the  $O(a^2 + ab + b^2)$ unique kernel entries. The $L$ factor can be further reduced by parallelizing the computation for multiple projected directions. 
Given $D_A$ has dimension $m$, steps~1-9 in Algorithm~\ref{alg:SW} only need to be precomputed once in $O(L|D_A| m  + L|D_A| \log |D_A|)$ time.

\begin{algorithm}
\caption{Sliced Wasserstein Distance Computation}\label{alg:SW}
\begin{flushleft}    
\textbf{Input}: Two dataset matrices $X_A, X_B$ with $m$ features/columns\\
\textbf{Parameter}: Number of projected directions $L$\\
\textbf{Output}: Sliced Wasserstein Distance $\SW(X_A$, $X_B)$ between $X_A$ and $X_B$, 
\end{flushleft}

\begin{algorithmic}[1]
\STATE $s \gets 0$
\FOR{$l = 1$ {\bfseries to} $L$}
\STATE Uniformly sample $\theta^{(l)}$ distributed on unit sphere 
    \STATE $\Pi_{\theta^{(l)}}(X_A) \gets X_A \theta^{(l)}$
    \STATE $\Pi_{\theta^{(l)}}(X_A) \gets \texttt{sort}(\Pi_{\theta^{(l)}}(X_A))$
    \STATE $\Pi_{\theta^{(l)}}(X_B) \gets X_B \theta^{(l)}$
    \STATE $\Pi_{\theta^{(l)}}(X_B) \gets \texttt{sort}(\Pi_{\theta^{(l)}}(X_B))$
    \STATE  $F_{X_A} \gets$ the empirical c.d.f.  with ${\Pi}_{\theta^{(l)}}(X_A)$
    \STATE  $F_{X_B} \gets$ the empirical c.d.f.  with ${\Pi}_{\theta^{(l)}}(X_B)$ 
    \STATE $s \gets s + \frac{1}{L}\left( \int_0^ 1 | F_{X_A}^{-1}(z)- F_{X_B}^{-1}(z)|^p dz \right) ^{1/p} $ \label{step:ot}
    
\ENDFOR
\STATE  \textbf{return} $s$
\end{algorithmic}
\end{algorithm}

\subsection{Supervised Learning}
In this section, we will additionally consider the target outputs $y_i$ for each owner $i$. Specifically, we will define a transformation $\mathcal{G}_{\eta}$ that will map the dataset (consists of both the input matrix and target outputs) to a common feature space. 

\begin{definition}
    Given the transformation $\mathcal{G}_{\eta}$, the \emph{supervised sliced Wasserstein} (SSW) distance between the empirical data distribution $D_A$ and $D_B$ is 
    \[
    \SSW_2^{2 \rho}(\mathbb{P}(D_A), \mathbb{P}(D_B); \mathcal{G}_{\eta}) = {\SW}_2^{2\rho}(\mathbb{P}(\mathcal{G}_{\eta}(D_A)), \mathbb{P}(\mathcal{G}_{\eta}(D_B)))\ .
    \]
\end{definition}

\textbf{Supervised regression.} Data valuation can be applied on regression problems, where the output $y_i$ is a vector of real values. We concatenate $y_i$ with $X_i$ and vary the weight on $y_i$ using a parameter $\eta$. Formally, let $\mathcal{G}_{\eta}(D_C) = \eta X_C \oplus (1-\eta) y_C$, where $\eta \in (0,1]$ is the scaling weight for the feature space.

\textbf{Supervised classification.} 
Data valuation is also often applied in supervised classification problems, where the output $y_i$ is a vector of discrete class labels.
At first glance, we can concatenate $y_i$ with $X_i$, however, how do we measure the distance between different labels such as `airplane', `bird', and `truck' in CIFAR-10? Inspired by \citet{alvarez2020otdd}, we quantify the distance between labels $\mathtt{y}^k$ and $\mathtt{y}^j$ as the sliced Wasserstein distance between the aggregated datasets with corresponding labels $\mathtt{y}^k$ and $\mathtt{y}^j$, i.e., $\SW(\mathbb{P}({x_i \mid (x_i, \mathtt{y}^k) \in D_N}), \mathbb{P}({x_i \mid (x_i, \mathtt{y}^j) \in D_N}))$. Thus, the 'bird' label is closer to `airplane' than `truck'.
We then use multi-dimensional scaling (MDS) \cite{pmlr-v139-demaine21a} to embed each class $\mathtt{y}^j$ as a vector $e(\mathtt{y}^j)$, preserving the distances between class labels. The function $e$ is also applied element-wise to embed the target outputs $y_C$. Formally, let $\mathcal{G}_{\eta}(D_C) = \eta X_C \oplus (1-\eta) e(y_C)$, where $\eta \in (0,1]$ are scaling weights for the feature and label spaces, respectively.

\begin{figure}[ht]
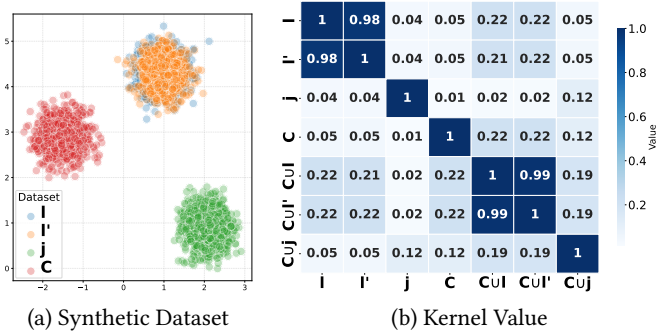

    \centering
    \begin{tabular}{@{}c@{\hspace{3mm}}c@{}}
        \includegraphics[width=0.4\columnwidth]{figures/circle.png} & \includegraphics[width=0.6\columnwidth]{figures/MDS_2_heatmap.png} \\ 
        (a) Synthetic Dataset & (b) Kernel Value
    \end{tabular}
    \caption{Datasets \textcolor{blue}{$l$} and \textcolor{orange}{$l'$} are similar while datasets \textcolor{blue}{$l$} and \textcolor{green}{$j$} are very different. The $\SSW$ kernel similarity $k(C \cup l, C \cup l') = .99$ is large as compared to the similarity $k(C \cup l, C \cup j) = 0.19$. 
    }
    \label{fig:intuition}
    \Description{Intuition of our Kernel in Gaussian Process}
\end{figure}

\begin{proposition}\label{prop:ssw}
    The kernel $k(A,B) = e^{-\gamma \SSW^{2\rho}_2(\mathbb{P}(D_A), \mathbb{P}(D_B); \mathcal{G}_{\eta})}$ is a valid kernel when $\gamma > 0$ and $\rho \in [0, 1]$ and $\eta > 0$.
\end{proposition}
The proof based on  Proposition~\ref{prop:sw} can be found in Appendix~\ref{app:proof_ssw}. The following toy example demonstrates why our chosen kernel makes intuitive sense. In Figure~\ref{fig:intuition}, as owners $l$ and $l'$ own similar data, the similarity $k(C \cup l, C \cup l')$ is large for any coalition $C$. In contrast, as owners $l$ and $j$ have very different data, the similarity $k(C \cup l, C \cup j)$ is always small. Thus, in a GP model, an observation of the utility of the coalition with $l$ would greatly reduce the uncertainty of that with $l'$ but not $j$.

\section{Semivalue Estimation}
\label{sec:semi}
We have defined the GP model and kernel to predict utilities for coalitions in $\cB$. Given the $a$ actual utility evaluations of $\cA$ (i.e., $\mathbf{u}_\cA$), we first learn the kernel hyperparameters (e.g., $\gamma$) and choose the norm of the $\SW$ distance and scaling weights $\eta$ for the feature and output spaces to maximize the log-likelihood. Then, we predict the posterior belief of the utilities $\hat{\mathbf{u}}_{\cB|\cA}$ of coalitions in $\cB$. How do we compute the semivalue, such as the Shapley value, based on the observed utilities $\mathbf{u}_\cA$ of coalitions in $\cA$ and the posterior belief $\hat{\mathbf{u}}_{\cB|\cA}$ of coalitions in $\cB$?

\begin{proposition}[Semivalue Prediction]\label{pos:shapley-dist} 
Let $\mathbf{w}^i_\cA$ and $\mathbf{w}^i_{\cB|\cA}$ denote the vectors containing the weights of all coalitions in $\cA$ and $\cB$ respectively.
The $j$-th entry of $\mathbf{w}^i_\cA$ corresponds to the weight of coalition $A_j$ in the computation of (i) the Shapley value $\phi_u(i)$ and (ii) the semivalue $\varphi_u(i)$, as defined in Equations~\ref{eq:shapley} and \ref{eq:semi}). 
When $i \notin A_j$, the $j$-th entry is given by (i) $-1/(n {\binom{n-1}{|A_j|}})$ or (ii) $-\omega_{|A_j|}$. When $i \in A_j$, the $j$-th entry is (i) $1/(n {\binom{n-1}{|A_j|-1}})$ or (ii) $\omega_{|A_j|-1}$.

The estimated Shapley value and semivalue for owner $i$, denoted as $\hat{\phi}_i$ and $\hat{\varphi}_i$, are given by the weighted sum $\mathbf{w}_{\cA}^i{}^\top \mathbf{u}_{\cA} + \mathbf{w}_{\cB}^i{}^\top \hat{\mathbf{u}}_{\cB|\cA}$ which follows the distribution
\[
\mathcal{N}( \mathbf{w}_{\cA}^i{}^\top \mathbf{u}_{\cA} + \mathbf{w}_{\cB}^i{}^\top \mathbb{E}[\hat{\mathbf{u}}_{\cB|\cA}], \mathbf{w}_{\cB}^i{}^\top \mathbb{V}[\hat{\mathbf{u}}_{\cB|\cA}] \mathbf{w}_{\cB}^i ).
\]
\end{proposition}

\begin{remark}[Monte Carlo Approximation]\label{rem:estimate}
    The weights $\mathbf{w}^i_\cA$ and $\mathbf{w}^i_{\cB|\cA}$ used to estimate $\hat{\phi}_i'$ can also correspond to the weights used in Monte Carlo estimates of Shapley value\cite{jia2019towards, kolpaczki2024without}, $\phi_i'$.
    In practice, when using Monte Carlo estimates, the sampled coalitions are collected and subsequently partitioned for actual evaluation and prediction, as illustrated in Figure~\ref{fig:framework}. 
\end{remark}

Proposition~\ref{pos:shapley-dist} considers only the uncertainty arising from the use of GP model predictions instead of actual utility evaluations (refer to \textbf{(ii)} in Section~\ref{sec:intro}). 
When Monte Carlo approximation is used, the total uncertainty in the estimate $\hat{\phi}_i'$ should also consider the uncertainty introduced by Monte Carlo sampling.
Formally, let $\sigma_{MC}$ denote the standard deviation introduced by Monte Carlo methods, such as those described by \citet{kolpaczki2024without}, which evaluates and utilizes only a subset of all coalitions (see \textbf{(i)} in Section~\ref{sec:intro}). Let $\sigma_{GP}$ represent the standard deviation associated with our GP model, defined as the square root of the variance in Proposition~\ref{pos:shapley-dist}. The total uncertainty $\mathbb{V}[\hat{\phi}_i']$ is upper bounded by $\sigma_{GP}^2 + \sigma_{MC}^2 + 2 \sigma_{GP} \sigma_{MC}$. 

\subsection{Active Querying to Accelerate Uncertainty Reduction}
We can reduce the variance $\sigma^2_{GP}$ associated with our GP model further by additionally evaluating $\bar{b}$ coalitions $\widebar{\cB} \subset \cB$. In particular, for each $B \in \widebar{\cB}$, the mediator trains a model on the aggregated data $D_B$ to evaluate $u(B)$ and only predicts the utility of the remaining coalitions $\cB \setminus \widebar{\cB}$.
Instead of randomly selecting  $\bar{b}$ coalitions, the mediator can actively select the $\bar{b}$ coalitions that lead to the largest reduction in the semivalue variance:
\begin{align}\label{eq:active}
    \widebar{\cB} &= \mathrm{argmax}_{\mathcal{C} \subseteq \cB, |\mathcal{C}| = \bar{b}}\ 
    \mathbf{w}^i_{\cB}{}^\top \cdot (\mathbb{V}[\mathbf{\hat{u}}_{\cB|\cA}] - \mathbb{V}[\mathbf{\hat{u}}_{\cB|\cA \oplus \mathcal{C}}]) \cdot \mathbf{w}^i_{\cB}. 
\end{align}
Here, $\oplus$ denote the concatenation operator and $\mathbb{V}[\mathbf{\hat{u}}_{\cB|\cA}]$ is the predictive variance of the utilities in $\cB$ given observed utilities from $\cA$.
As the weighted variance reduction function is often monotone submodular \cite{das2008algorithms}, Equation~\eqref{eq:active} can be maximized by the greedy algorithm in Algorithm~\ref{alg:active}. 

\begin{algorithm}
\caption{Greedy Active Selection with Efficient Inverse Update}\label{alg:active}
\begin{flushleft}
\textbf{Input}: Evaluated coalitions and utilities, $(\mathcal{A}, \mathbf{u}_{\mathcal{A}})$; unevaluated coalitions, $\mathcal{B}$; kernel function $k$; semivalue weight vector $\mathbf{w}_{\mathcal{B}}^i$ \\
\textbf{Parameter}: Number of additional evaluations $\bar{b}$ \\
\textbf{Output}: Selected coalitions $\bar{\mathcal{B}}$ for additional evaluations
\end{flushleft}
\begin{algorithmic}[1]
    \STATE Initialize: $\widebar{\cB} \gets ()$, $\mathbf{K}^{-1} \gets (K_{\mathcal{A}, \mathcal{A}} + \sigma^2 \mathbf{I})^{-1}$
    \FOR{$j = 1$ {\bfseries to} $\bar{b}$}
        \STATE Set $\mathtt{max\_VR} \gets 0$, $G^* \gets \textit{null}$
        \FOR{$G \in \mathcal{B} \setminus \widebar{\cB}$}
            \STATE Let $\mathcal{C} \gets \widebar{\cB} \oplus (G,)$
            \STATE Incrementally update $\mathbf{K}^{-1}$ using the previous inverse and $k(G, \mathcal{A} \oplus \widebar{\cB})$ (as described in Appendix~\ref{appdix:algo2})
            \STATE Compute the variance reduction:
            \[
            VR \gets (\mathbf{w}_{\mathcal{B}}^i)^\top \cdot
            \mathbf{K}_{\mathcal{B}, \mathcal{A} \oplus \mathcal{C}}
            \mathbf{K}^{-1}
            \mathbf{K}_{\mathcal{A} \oplus \mathcal{C}, \mathcal{B}}
            \cdot
            \mathbf{w}_{\mathcal{B}}^i
            \]
            \IF{$VR \geq \mathtt{max\_VR}$}
                \STATE Update $\mathtt{max\_VR} \gets VR$, $G^* \gets G$
            \ENDIF
        \ENDFOR
        \STATE Add the selected coalition: $\widebar{\cB} \gets \widebar{\cB} \oplus (G^*,)$
    \ENDFOR
    \STATE \textbf{return} $\widebar{\cB}$
\end{algorithmic}
\end{algorithm}

\section{Experiments}\label{sec:exp}
We conduct experiments across several datasets to evaluate the effectiveness of our methods. For classification tasks, we utilize (a) the Moon dataset \cite{scikit_learn}, (b) the MNIST dataset \cite{deng2012mnist}, (c) the CIFAR-10 dataset \cite{Krizhevsky09}, and (d) the IMDb dataset \cite{maas2011learning}. For regression tasks, we employ (e) the California Housing dataset (CaliH) \cite{SPL97_pace1997sparse}, which provides real-world housing data. We use accuracy as the utility function for classification tasks and the $R^2$ score for regression tasks, defined as $R^2 = 1 - \frac{SS_{\text{res}}}{SS_{\text{tot}}} = 1 - \frac{\sum_{i=1}^{n} (y_i - \hat{y}_i)^2}{\sum_{i=1}^{n} (y_i - \overline{y})^2}$. Finally, in our experiments, we employ the Shapley value, the most widely used semivalue.

Our experiments aim to achieve three primary objectives: (i) to investigate how factors such as the kernel and the number of randomly and actively selected coalitions affect the performance and computation time of our framework in computing Shapley values; (ii) to demonstrate the application of our framework in data valuation, particularly for classification tasks; and (iii) to highlight the advantages of uncertainty quantification.

\textbf{Baseline:} We evaluate three baselines to compare with our proposed approach. The first baseline, denoted as \texttt{OTDD}, is based on the label-feature distance concept from \citet{alvarez2020otdd}. Although this baseline employs the optimal transport distance between feature-label pairs, it is important to note that the exponential kernel derived from \texttt{OTDD} is not valid, unlike the kernel we propose. We summarize the computational complexity of each distance metric in Appendix \ref{sec:A:compare}, Table~\ref{tab:dis_compare}. For this baseline, we implement the Sliced Wasserstein distance as the label distance for efficiency.\footnote{\citet{alvarez2020otdd} originally defined the distance as $\left(\|x - x'\|^p + W_p^p\right)^{1/p}$, where $W_p^p$ is the $p$-Wasserstein distance between label distributions. In our implementation, we use the Sliced Wasserstein (SW) distance for improved efficiency.} The second and third baselines, inspired by \citet{wang2021predict}, represent each dataset $D_i$ with a binary indicator vector (01 encoding) $b_i \in \{0,1\}^n$, indicating the indices present for each data owner $i$. These methods are referred to as \texttt{GP-binary} and \texttt{NN-binary}, employing Gaussian Process Regression and Neural Networks, respectively. Finally, for our SSW kernel, we set $\eta=0.5$.

\subsection{Coalition Utility Prediction}\label{sec:cup}
This section empirically evaluates the utility prediction performance of our proposed kernel $\SSW$ and the baseline methods. We compare the computation time between our $\texttt{DUPRE}$ framework and an exhaustive evaluation of all possible coalitions. Our experiments include two settings: one where $\alpha$ randomly selected coalitions' utilities are evaluated by model training and another (with -a suffix) where half of the coalitions are randomly selected and the other $\alpha/2$ coalitions are actively selected as described in Algorithm~\ref{alg:active} and Figure~\ref{fig:framework}. After evaluating these $\alpha$ coalitions, we train our $\texttt{DUPRE}$ framework on their utilities and then estimate the utilities of all remaining coalitions. 

To assess the effectiveness of our methods, we compute the mean and standard deviation of the mean squared error (MSE) between our predictions and the actual utilities over ten runs, each using a different set of randomly evaluated coalitions (with random seeds from 0 to 9). We also calculate the Pearson correlation coefficient \cite{kirchPearson2008} to evaluate the correlation between predicted and actual utilities and use Kendall's tau metric \cite{kendall1938new} to assess the ranking order of predicted and actual Shapley values, which is critical for understanding a data owner’s contribution.

\begin{figure}[ht]
\includegraphics[width=1\columnwidth]{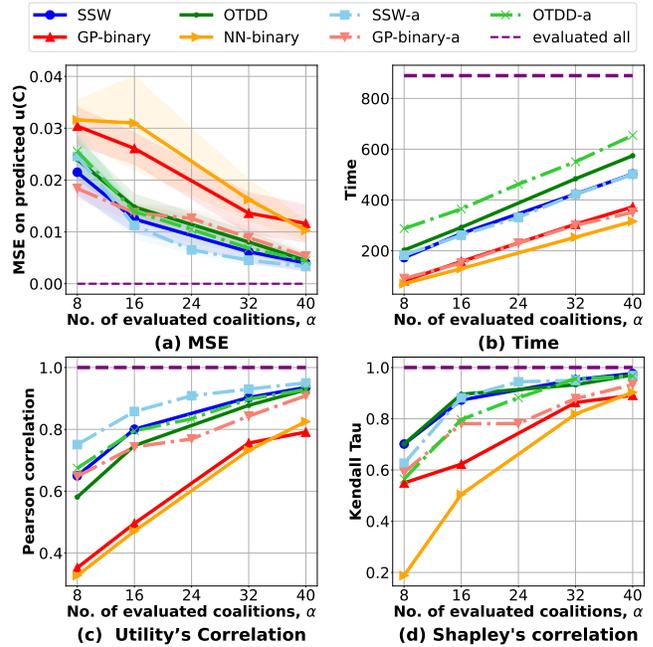} 
    \caption{A comparison of the quality of the utility predictions and time taken for various methods on the MNIST dataset with $6$ owners.
    The \texttt{-a} suffix indicates that we have actively selected 50\% of the coalitions to accelerate uncertainty reduction.
    }
    \label{fig:up_mnist_6}
    \Description{Comparison of methods on MNIST with six data owners, showing MSE, correlation, and computation time for coalition utility prediction.}
\end{figure}

We first consider predicting the validation accuracy on the MNIST dataset, using a neural network as the ML model for evaluation. Each of the $6$ data owners holds a distinct subset of digit labels: $\{1, 4, 5, 7, 8\}, \{2\}, \{9\}, \{6\}, \{0\}, \{3\}$. As shown in Figure~\ref{fig:up_mnist_6}, our kernel ($\SSW$) outperforms the other baselines, evidenced by a lower mean squared error (MSE) in (a) and higher Pearson correlation coefficients in (c). While the active selection process improves performance, it also increases computation time. Nonetheless, our \texttt{DUPRE} framework remains faster than an exhaustive actual evaluation of all possible coalitions. Moreover, our kernel provides a superior ranking of Shapley values than other kernels, as indicated by the higher Kendall's tau coefficients in (d).

\begin{figure}[ht]
\includegraphics[width=1\columnwidth]{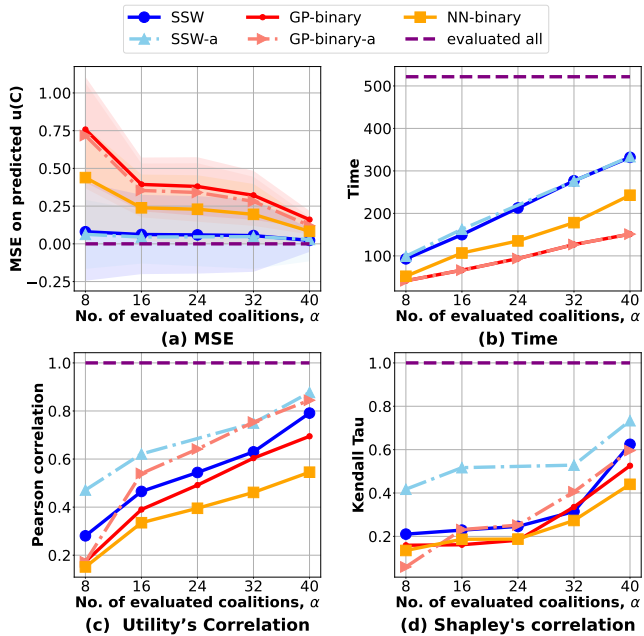}
    \caption{
    A comparison of the quality of the utility predictions and time taken for various methods on the CaliH dataset with $6$ owners.
    The \texttt{-a} suffix indicates that we have actively selected 50\% of the coalitions to accelerate uncertainty reduction.}
    \label{fig:up_calih_6}
    \Description{Comparison of methods on California Housing with six data owners, showing MSE, correlation, and computation time for coalition utility prediction.}
\end{figure}
We repeat this experiment on the regression dataset, CaliH, as illustrated in Figure~\ref{fig:up_calih_6}, and use Multi-Layer Perceptron (MLP) as the ML model. The kernel based on \texttt{OTDD} is not applicable here, as it requires classification labels. Once again, our proposed kernel results in lower MSE and higher correlations than the baselines. 

\subsection{Evaluating the Quality of Shapley Value Predictions}
\label{sec:quality_shapley}
In this section, we explore two approaches for computing the Shapley value: predicting the utilities of all coalitions to estimate exact Shapley values and predicting only the utilities of a subset to estimate approximate Shapley values.
\subsubsection{Exact Shapley Value Estimation}
\label{sec:exp_exact_shap}
We evaluated our framework on the CIFAR-10 and CaliH datasets with $8$ data owners. For CIFAR-10, the utility function is the accuracy of the trained ResNet model. In Table~\ref{tab:compare_corr}, the first segment computes the Shapley value based on the predicted utilities from a GP model that is trained on the utility of $100$ actually evaluated coalitions.
In contrast, the GP model in the second segment is additionally trained on the utility of $10$ more coalitions, randomly selected or selected by Algorithm~\ref{alg:active}.

\begin{table*}[ht]
\caption{A comparison of the quality of exact Shapley value approximation and the time taken on the CIFAR-10 and CaliH datasets with $8$ data owners. A higher correlation is preferred. }
    \centering
    \label{tab:compare_corr}
    \begin{tabular}{l|ccc|ccc}
        \toprule
        \multirow{2}{*}{\textbf{Method}} & \multicolumn{3}{c}{\textbf{CIFAR-10}} & \multicolumn{3}{c}{\textbf{CaliH}} \\ 
        & \textbf{Pearson} & \textbf{Kendall Tau} & \textbf{Time (s)}  & \textbf{Pearson} & \textbf{Kendall Tau} & \textbf{Time (s)} \\ 
        \midrule
        $\texttt{SSW}$ & \textbf{0.901} $\pm$ 0.07 & \textbf{0.664} $\pm$ 0.152 & 5893 $\pm$ 1375 &  \textbf{0.775} $\pm$ 0.181 & \textbf{0.6714} $\pm$ 0.18 & 543 $\pm$ 10 \\ 
        $\texttt{OTDD}$ & 0.640 $\pm$ 0.09 & 0.523 $\pm$ 0.21 & 8056 $\pm$ 2141 & - & - & - \\ 
        $\texttt{GP-binary}$ & 0.785 $\pm$ 0.006 & 0.565 $\pm$ 0.1 & 4120 $\pm$ 593 & 0.528 $\pm$ 0.134 & 0.593 $\pm$ 0.129 & 264 $\pm$ 35 \\ 
        $\texttt{NN-binary}$ & 0.612 $\pm$ 0.01 & 0.544 $\pm$ 0.12 & 2541 $\pm$ 256 & 0.579 $\pm$ 0.154 & 0.602 $\pm$ 0.163 & 351 $\pm$ 45\\ 
        $\texttt{LAVA}$ & -0.0785 $\pm$ 0.0405 & -0.3045 $\pm$ 0.03 & 5580 $\pm$ 394 & 0.1644 $\pm$ 0.226 & 0.107 $\pm$ 0.15 & 50 $\pm$ 10 \\ \hline
        \textbf{Evaluate $10$ additional coalitions} \\
         $\texttt{SSW}$ - random & 0.911 $\pm$ 0.04 & 0.674  $\pm$ 0.132  & 6137 $\pm$ 1098  &    0.805 $\pm$ 0.155 & 0.7124 $\pm$ 0.21 & 585 $\pm$ 20\\  
	$\texttt{SSW}$ - active & \textbf{0.934} $\pm$ 0.07 & \textbf{0.677}  $\pm$ 0.126  & 6317 $\pm$ 1567  &    \textbf{0.831} $\pm$ 0.165 & \textbf{0.7624} $\pm$ 0.24 &     627 $\pm$ 23\\ 
        $\texttt{GP-binary}$ - active &   0.855 $\pm$ 0.006 & 0.615 $\pm$ 0.15 & 4320 $\pm$ 635 & 0.655 $\pm$ 0.1 & 0.653 $\pm$ 0.120 & 388 $\pm$ 40 \\\hline 
        \textbf{Evaluate all coalitions} &   - & - & 12458   & - & -  & 950   \\
        \bottomrule

    \end{tabular}
\end{table*}

We use Pearson and Kendall's tau correlation coefficients as evaluation metrics to assess the agreement between our predicted and the exact Shapley values, i.e., $(\hat{\phi}_i)_{i \in N}$ and $(\phi_i)_{i \in N}$. We also compare against \texttt{LAVA} \cite{just2023lava}, a model-agnostic method that estimates Shapley values based on the distance between datasets and the task dataset, without requiring ML model training. Our method, \texttt{SSW}, produces estimates that are more correlated with the exact Shapley values, as evidenced by higher Pearson and Kendall's tau coefficients as compared to other kernels. In contrast, the \texttt{LAVA} method results in the lowest correlations. 
Additionally, we observe that active selection, which incurs a slightly higher computation cost, improves the correlation more than random selection. 

\subsubsection{Approximate Shapley Value Estimation}\label{sec:est_sampling_shapley}
Next, we evaluate our approach using the MNIST dataset distributed among $10$ data owners. We compute the approximate Shapley values using permutation sampling, as outlined by \citeauthor{castro2009polynomial} \cite{castro2009polynomial}. We limit the number of permutation samples considered by the total number of evaluated and predicted coalitions. Refer to Appendix~\ref{appdix:samp_algo} for more details.

In our experiments, we actually evaluate the utilities of $512$ coalitions by ML model training and use these utilities to train our GP model. Then, we either consider evaluating the utilities of another additional $250$ coalitions (purple line) or predicting the utilities of $\beta$ additional coalitions ($\beta \in [0, 250]$) using the GP model with different kernels. The key objective of the experiment is to determine if these predicted utilities can effectively substitute for actual utility evaluations.

\begin{figure}[ht]
\includegraphics[width=1.\columnwidth]{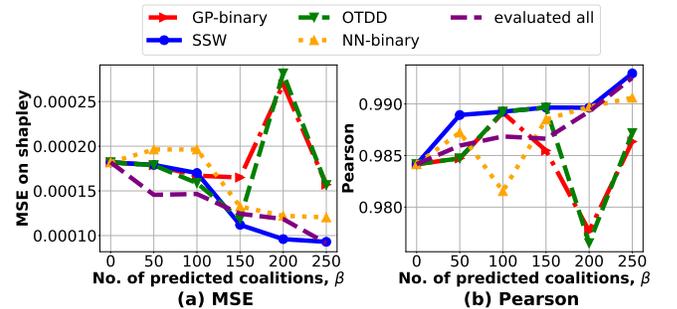} 
\caption{
A comparison of the quality of the approximate Shapley value predictions for various methods on the MNIST dataset. The approximate Shapley value is computed using $512$ actual utility evaluations and an additional $\beta$ actual/predicted coalition utilities. 
}
    \label{fig:per}
    \Description{Comparison of approximate Shapley value predictions for various methods on MNIST, computed using 512 actual utility evaluations plus additional actual/predicted coalition utilities.}
\end{figure}

In Figure~\ref{fig:per}, the x-axis represents the number of additional coalitions used to compute the approximate Shapley value $\phi_i'$. We measure the mean squared error (MSE) between the estimated approximate Shapley value $\hat{\phi}_i'$ and the exact Shapley value $\phi_i$ computed using Equation~\ref{eq:shapley} as well as the Pearson correlation between the estimated approximate Shapley values and exact Shapley values across data owners. The purple lines exhibit a strictly decreasing MSE and a strictly increasing Pearson correlation as more coalitions are evaluated. Similarly, for our SSW kernel, both the MSE and Pearson correlation improve as more coalitions are predicted. Given that the performance of the SSW kernel mirrors that of the actual evaluations, it is a suitable substitute. In contrast, the other baselines do not exhibit the same trend; for instance, with NN-binary, increasing the number of predicted utilities can worsen both the MSE and Pearson correlation.

\subsection{Benefits of Uncertainty Quantification}
\label{sec:benfun}
\begin{figure}[ht]
\includegraphics[width=1.\columnwidth]{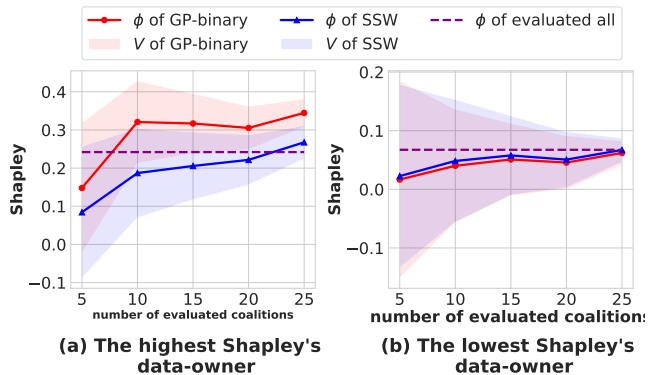} 
\caption{Plot of the GP model's predictive mean and standard deviation (shaded region around the line) of Shapley values for five data owners.
}
    \label{fig:bef}
    \Description{Uncertainty quantification among five MNIST data owners. 
    }
\end{figure}

In this experiment, we examine the uncertainty of the predicted Shapley values of the MNIST classification task where the dataset is split among 5 data owners. We specifically focus on the data owner with (a) the highest contribution and (b) the lowest contribution. Since our kernel $\SSW$ outperforms \texttt{OTDD}, we consider only \texttt{GP-binary} as the baseline. We compute the Shapley values and their variances using the formula in Section~\ref{sec:pf}. The result is illustrated in Figure~\ref{fig:bef}.

As the number of evaluated coalitions increases, the predicted Shapley value gets closer to the actual Shapley value and the variance decreases. For our $\SSW$ kernel, the actual Shapley value always lie within the shaded region, suggesting our uncertainty quantification is well-calibrated.

\subsection{Further Analysis} 
\label{sec:further-analysis}
In this section, we present additional experiments to analyze and stress-test our framework. 
First, we perform an ablation study on the parameter $\eta$ (Section~\ref{subsec:ablation}), 
which controls the relative importance of label information in our kernel. Next, we demonstrate how to handle more complex datasets like IMDb (Section~\ref{subsec:nlp-datasets}). Finally, we show the robustness of our method in a heavily imbalanced and heterogeneous setting using the IMDb dataset (Section~\ref{subsec:imdb-robustness}).

\subsubsection{Effect of the label-weight parameter $\eta$}
\label{subsec:ablation}
We perform an ablation study to understand the influence of different $\eta$ (i.e., different weight of the output label) affects the mean squared error (MSE) of the utility and Shapley value predictions. This experiment uses a classification task on the synthetic Moon dataset with $6$ data owners.

\begin{figure}[ht]
\includegraphics[width=1.\columnwidth]{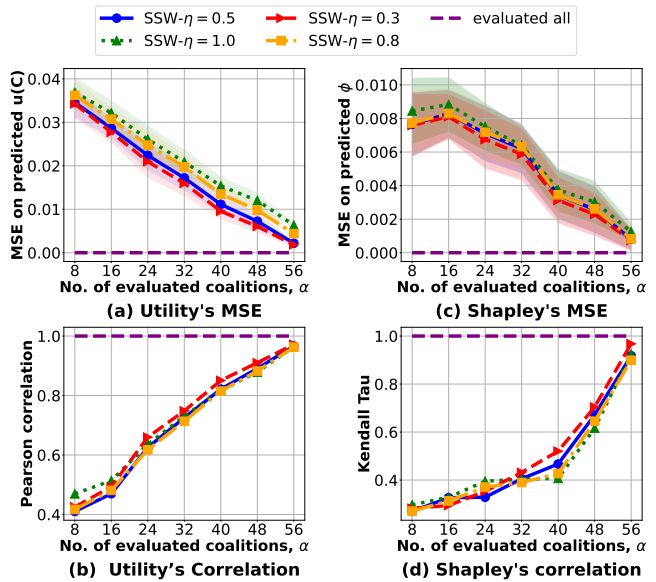} 
\caption{
    A comparison of the quality of the utility and Shapley value predictions for various $\eta$ values on the Moon dataset with $6$ data-owners. }
    \label{fig:abamoon}
    \Description{Different label-weight parameter on the Moon dataset}
\end{figure}

A smaller $\eta$ value means that the label information has a greater influence on the dataset distance, allowing the GP model to better capture label-dependent patterns. As illustrated in Figure~\ref{fig:abamoon}, $\eta = 0.3$ results in the lowest MSE and highest Pearson correlation, indicating that assigning more weight to label information leads to improved predictive performance.

\subsubsection{Evaluation on an unstructured dataset with 10 data owners}
\label{subsec:nlp-datasets}
\begin{table*}[ht!]
\centering
\caption{
\textbf{A comparison of the quality of utility predictions and Shapley value predictions for various methods on the IMDb dataset with $10$ owners.}
\textbf{Setup~1 (similar to Section \ref{sec:cup})} compares the actual utility $u(C)$ with GP predicted expected utility $\hat{u}(C)$ of various coalition $C$.
\textbf{Setup~2 (similar to Section \ref{sec:est_sampling_shapley})} compares the actual Shapley value $\phi_i$ and the predicted Shapley value $(\hat{\phi}_i)$ for various owner $i$. 
The results are the mean $\pm$ std.\ over 5 runs.
Lower MSE and higher correlation are preferred.
}
\label{tab:imdb-combined-main}
\begin{tabular}{l|ccc|cc}
\toprule
& \multicolumn{3}{c|}{\textbf{Setup 1 (Section \ref{sec:cup})}} 
& \multicolumn{2}{c}{\textbf{Setup 2 (Section \ref{sec:est_sampling_shapley}) }} \\
\textbf{Method} 
& $\text{MSE}(u(C), \hat{u}(C))$ 
& $\text{Pearson}\bigl(u(C), \hat{u}(C) \bigr)$
& $\text{Shapley Corr.}$
& $\text{MSE}((\phi_i)_{i \in N}, (\hat{\phi}_i)_{i \in N})$
& $\text{Pearson}((\phi_i)_{i \in N}, (\hat{\phi}_i)_{i \in N})$ \\
\midrule
\texttt{SSW (Ours)} 
& $\mathbf{8.6\times10^{-6}\pm0.008}$ 
& $\mathbf{0.60\pm0.13}$ 
& $\mathbf{0.75\pm0.21}$
& $\mathbf{0.00022\pm0.00016}$ 
& $\mathbf{0.59\pm0.22}$\\
\texttt{GP-binary}  
& $1.8\times10^{-5}\pm0.0042$ 
& ${0.466\pm0.17}$ 
& $0.66\pm0.26$
& $0.00029\pm0.00014$
& $0.50\pm0.24$ \\
\texttt{NN-binary}  
& $0.006\pm0.008$ 
& $-0.199\pm0.07$ 
& $-0.137\pm0.35$
& $0.0001\pm0.00015$
& $0.52\pm0.025$ \\
\bottomrule
\end{tabular}
\end{table*}
We now evaluate our framework on the IMDb dataset, which comprises $50000$ movie reviews labeled as either positive or negative. Following the OpenDataVal benchmark \cite{opendataval2023}, we use DistilBERT \cite{sanh2019distilbert} embeddings for each review. We split the dataset among 10 data owners and consider two experimental settings as in Sections~\ref{sec:cup} and \ref{sec:est_sampling_shapley}). For the former, we use 256 evaluated coalitions. For the latter, we use 512 evaluated coalitions and 100 predicted coalitions.

Table~\ref{tab:imdb-combined-main} shows that our method $\SSW$ always achieves lower MSE and higher correlation than 
\texttt{GP-binary} and \texttt{NN-binary}, across five runs where the evaluated and predicted coalitions are randomly varied. In Appendix~\ref{appendix:SST-2}, we further validate our framework on the \emph{Stanford Sentiment Treebank (SST-2)} \cite{socher-etal-2013-recursive} dataset. We also observe that $\SSW$ consistently outperforms the baselines across different text-based tasks and pre-trained embeddings.

\subsubsection{Robustness to Heterogeneous Data Size and Distribution}
\balance
While the earlier experiments already account for varying dataset sizes and distributions (see Appendix~\ref{appendix:exp}, Table~\ref{tab:datasize_per_owner}), we further validate our framework under more extreme heterogeneity using the IMDb dataset. Specifically, we consider $20$ data owners, where the $j$-th data owner has $100j$ data points. Additionally, owners $j=1...5$ hold only data of the positive class, while owners $j=6...10$
hold only data of the negative class. The remaining $29000$ data points are used as the validation set. 
\label{subsec:imdb-robustness}
\begin{table}[ht!]
    \centering
    \caption{
    A comparison of the quality of the Shapley value predictions for various methods on the IMDb dataset with $20$ heterogeneous data owners. The Shapley value are computed based on $5000$ actual utility evaluations and $20000$ predicted utility evaluations.
    }
    \label{tab:imdb}
    \begin{tabular}{lcc}
    \toprule
    \textbf{Method} & $\text{MSE}$ & $\text{Pearson}$\\
    \midrule
    \texttt{SSW (Ours)}        &  ${9.60\times10^{-5}}$ &  $\mathbf{0.736}$\\
    \texttt{GP-binary }                 & $\mathbf{9.48\times10^{-5}}$ & $0.710$\\
    \texttt{NN-binary}                  & $1.00\times10^{-4}$ & $0.652$ \\
    \bottomrule
    \end{tabular}
\end{table}

We consider estimating the Shapley value with $25000$ coalitions ($\sim1800$ permutations, see Appendix~\ref{appdix:samp_algo}). Out of these $25000$ coalitions, we actually evaluate the utilities of $5000$ random coalitions and predict the utilities of the remaining coalitions. 
In Table~\ref{tab:imdb}, we observe that despite the large variation in dataset sizes and distributions, $\SSW$ achieves the strongest correlation (0.736) and low MSE ($9.60\times10^{-5}$). 
Additionally, we observe that the MSE does not increase when the owners have more data points. The Pearson correlation between dataset size and MSE for $\SSW$ is low and only \textbf{0.241}. 

\section{Related Work}
Our work is complementary to related works on \emph{data valuation} that propose new data utility functions (such as data volume \cite{NEURIPS2021_59a3adea} and information gain \cite{sim2020cml}) and strategies (such as the Shapley value \cite{ghorbani2019data, Kwon2021betashapley, yan2021core}, the Banzhaf value \cite{wang2023data} and Least Core \cite{yan2021core}). \texttt{DUPRE} can be used to efficiently predict the utilities of any data utility function needed in the data valuation strategies.

Our work is also complementary to \emph{semivalue approximation techniques} that reduce the number of coalitions to evaluate such as permutation sampling \cite{castro2009polynomial}, stratified sampling \cite{Maleki.2013}, structured sampling \citep{vanCampen.2018}, or approximating Shapley without marginal contribution \cite{kolpaczki2024without}. Instead of evaluating all the sampled coalitions' utilities by training a model, we propose evaluating a subset and predicting the remaining utilities using a GP model. Our work offers an alternative to methods that reduce the cost per evaluation, such as TMC-Shapley, Gradient Shapley \cite{ghorbani2019data} and the influence function heuristic used by \citet{jia2019towards}.

Our work can be extended to make use of other \emph{dataset distances}, such as optimal transport dataset distance (OTDD) \cite{alvarez2020otdd}, if they satisfy the valid properties of a kernel. There are also other data valuation works that have used the SW distance or aim to reduce the cost of each utility function but they differ in their application. In data valuation works, \citet{just2023lava,kessler2024sava} have also defined their data utility function based on the optimal transport distance between training data subsets and validation data. However, our purpose of considering dataset distances is different.

\section{Conclusion}
In this paper, we introduce \texttt{DUPRE}, a novel framework that complements existing sampling-based approximation methods to further boost the efficiency of computing CGT-based data valuation. We design a valid kernel based on the sliced Wasserstein distance and adapt the distance to consider the target outputs in supervised learning. As our kernel can encode prior knowledge of similarities between different data subsets, our GP model outperforms other approaches in our experiments.

While \texttt{DUPRE} demonstrates strong empirical performance, its predictions are not guaranteed to be accurate for every dataset or utility function. We recommend using a validation set of coalitions and utilities to continually assess and improve its predictions. 
Future work can consider other applications of data utility prediction, such as in online data valuation scenarios where new data owners frequently join or leave the collaboration.

\begin{acks}
This research is supported by the National Research Foundation Singapore and DSO National Laboratories under the AI Singapore Program (AISG Award No: AISG2-RP-2020-018).
\end{acks}
\bibliographystyle{ACM-Reference-Format}
\bibliography{refs}

\ifshowappendix
\appendix
\onecolumn 
\section{Background}
\subsection{Gaussian Process Regression}
\label{appendix:gp_background}
A Gaussian Process (GP)~\cite{williams_gaussian_1996} is a collection of random variables in which any finite subset follows a joint Gaussian distribution. Formally, a GP is specified by its mean function $m(x)$ and covariance (kernel) function $k(x, x')$. For any set of $p$ input points $x_1, \dots, x_p$, the joint distribution of the corresponding function values is
\begin{equation}
    f(x_1), \dots, f(x_p) 
    \;\sim\;
    \mathcal{N}(\mathbf{m}, \mathbf{K}),
\end{equation}
where $\mathbf{m}$ is the mean vector with entries $m(x_i)$, and $\mathbf{K}$ is the covariance matrix whose $(i,j)$-th entry is $k(x_i, x_j)$. 
Given training data $(X, y)$ and a new input $x_*$, the predictive distribution for $f(x_*)$ is:
\begin{equation}
    f(x_*) \;\bigm|\; X, y, x_* 
    \;\sim\; 
    \mathcal{N}(\mu_*, \sigma_*^2),
\end{equation}
with
\[
    \mu_* 
    = \mathbf{k}_*^T \bigl(\mathbf{K} + \sigma_n^2 \mathbf{I}\bigr)^{-1} y,
    \quad
    \sigma_*^2
    = k(x_*, x_*) 
    \;-\; \mathbf{k}_*^T\,\bigl(\mathbf{K} + \sigma_n^2 \mathbf{I}\bigr)^{-1}\,\mathbf{k}_*.
\]
Here, $\mathbf{k}_*$ is the vector of covariances between the training inputs $X$ and the new input $x_*$, $\sigma_n^2$ is the noise variance, and $\mathbf{I}$ is the identity matrix.
Key Features of Gaussian Processes:
\begin{itemize}
    \item \textbf{Flexibility:}
    GPs are non-parametric, allowing them to adapt to diverse datasets without being constrained by a fixed functional form.

    \item \textbf{Uncertainty Estimation:}
    GPs naturally quantify how certain (or uncertain) they are about each prediction, unlike many methods that provide only a point estimate.

    \item \textbf{Kernel Choice:}
    The kernel (covariance) function encodes assumptions about the underlying function. Popular options include the Radial Basis Function (RBF) and Mat\'ern kernels. Kernel parameters (e.g., length scale) are commonly learned by maximizing the log marginal likelihood.
\end{itemize}

\subsection{Optimal Transport Dataset Distance}
\label{appendix:otdd_background}
The concept of \textbf{Optimal Transport (OT)} originates from 18th-century France, where mathematician Gaspard Monge sought the most efficient method to transport soil. Consider a space $\mathcal{X}$ equipped with a probability measure $\mathcal{P}(\mathcal{X})$. For a joint measure $\pi \in \mathcal{P}(\mathcal{X} \times \mathcal{X})$, the marginals are denoted by $P_{1\#}\pi$ and $P_{2\#}\pi$, corresponding to the projection maps $P_1(x, x') = x$ and $P_2(x, x') = x'$, respectively.
Given two probability measures $\alpha, \beta \in \mathcal{P}(\mathcal{X})$, the Kantorovich formulation of the optimal transport problem is defined as:
\begin{equation}\label{eq:ot}
    \text{OT}(\alpha, \beta) \triangleq \min_{\pi \in \Pi(\alpha, \beta)} \int_{\mathcal{X} \times \mathcal{X}} c(x, x') \, d\pi(x, x'),
\end{equation}
where $\Pi(\alpha, \beta)$ denotes the set of all couplings between $\alpha$ and $\beta$:
\[
\Pi(\alpha, \beta) = \left\{ \pi \in \mathcal{P}(\mathcal{X} \times \mathcal{X}) \mid P_{1\#}\pi = \alpha, \, P_{2\#}\pi = \beta \right\},
\]
and $c: \mathcal{X} \times \mathcal{X} \rightarrow \mathbb{R}^+$ is a cost function measuring the "transportation cost" between elements $x$ and $x'$.

\noindent A particularly important instance of optimal transport is the \textbf{squared Wasserstein distance}. When the cost function is chosen as the squared Euclidean distance $c(x, x') = \|x - x'\|^2$,
the Kantorovich formulation in Equation~\eqref{eq:ot} specializes to
\[
W_2^2(\alpha, \beta) \triangleq \min_{\pi \in \Pi(\alpha, \beta)} \int_{\mathcal{X} \times \mathcal{X}} \|x - x'\|^2 \, d\pi(x, x').
\]

\noindent Building upon OT, \citet{alvarez2020otdd} introduced the \emph{Optimal Transport Dataset Distance} (OTDD), which incorporates both features and labels of datasets. The OTDD between two datasets $D_A$ and $D_B$ is defined as:
\begin{align*}
\texttt{OTDD}(D_A, D_B) &= \min_{\pi \in \Pi(\mathbb{P}(A), \mathbb{P}(B))} \int d_Z\big( (x, y), (x', y') \big)^p \, d\pi\big( (x, y), (x', y') \big) \\
&= \min_{\pi \in \Pi(\mathbb{P}(A), \mathbb{P}(B))} \int \left[ d_{\mathcal{X}}(x, x')^p + d_{\mathcal{Y}}(y, y')^p \right] \, d\pi\big( (x, y), (x', y') \big),
\end{align*}
where $d_Z$ is a metric on the product space $Z = \mathcal{X} \times \mathcal{Y}$, $d_{\mathcal{X}}$ and $d_{\mathcal{Y}}$ are metrics on the feature space $\mathcal{X}$ and the label space $\mathcal{Y}$, respectively, and $p \geq 1$.

Let $D_A^c = \{ x \in D_A \mid y = c \}$ denote the subset of features in dataset $D_A$ associated with label $c$. Similarly, let $\mathbb{P}(D_A^y)$ represent the empirical distribution of features in $D_A$ conditioned on label $y$. With these definitions, the OTDD can be expressed as:
\[
\texttt{OTDD}(D_A, D_B) = \int \left( \| x - x' \|^p + W_p^p\big( \mathbb{P}(D_A^y), \mathbb{P}(D_B^{y'}) \big) \right)^{1/p} \, d\pi\big( (x, y), (x', y') \big),
\]
where $W_p^p\big( \mathbb{P}(D_A^y), \mathbb{P}(D_B^{y'}) \big)$ is the $p$-th power of the $p$-Wasserstein distance between the conditional feature distributions given labels $y$ and $y'$. For computational efficiency, we approximate the Wasserstein distance using the Sliced Wasserstein distance ($\SW_1$), thus setting:
\[
W_p^p\big( \mathbb{P}(D_A^y), \mathbb{P}(D_B^{y'}) \big) = \SW_1\big( \mathbb{P}(D_A^y), \mathbb{P}(D_B^{y'}) \big).
\]
This approximation enhances efficiency while maintaining a meaningful measure of the distance between datasets in terms of both their features and label distributions.

\section{Problem Formulation}\label{appendix:pos}

\begin{definition}[Valid kernel]
A valid kernel is a function $k(\mathbf{x},\mathbf{x}')$ that corresponds to a scalar (inner) product in some (perhaps infinite dimensional) feature space.
\[
k(\mathbf{x}, \mathbf{x'}) = \Phi(\mathbf{x}){}^{\top}\Phi(\mathbf{x}).
\]
One consequence of this is that kernel functions must be symmetric, since $\Phi(\mathbf{x}){}^{\top}\Phi(\mathbf{y}) =  \Phi(\mathbf{y}){}^{\top}\Phi(\mathbf{x})$.
\end{definition}
\begin{definition}[Positive Semi-Definiteness]
A function $ k: X \times X \rightarrow \mathbb{R} $ is called a \emph{positive semi definite} if for any finite set $ \{ x_1, x_2, \dots, x_n \}  \subset X $ and any real numbers $ c_1, c_2, \dots, c_n $, it holds that:
  \[
  \sum_{i=1}^{n} \sum_{j=1}^{n} c_i c_j\, k(x_i, x_j) \geq 0.
  \]
\end{definition}

Our positive semidefinite kernel definition is also referred to as positive definite kernel in \cite{Kanagawa2018GaussianPA}.


In Gaussian Process Regression, the choice of the kernel is important because it encodes our assumptions about the function we aim to learn. A positive definite kernel ensures that the covariance matrices constructed during GPR are nondegenerate—that is, they have strictly positive eigenvalues and are invertible. This guarantees that the mathematical and computational procedures involved during training and prediction are well-defined. Without positive definiteness, the covariance matrix could be singular or ill-conditioned, leading to numerical instability and unreliable or undefined results.



\begin{proposition}(Proposition 14 in \cite{meunier2022slicedw} or Proposition 2.1 in \cite{Haasdonk})\label{pos:pd_kernel}.

Let $ M$ be a set and $d$ be a pseudo-distance on $M$. The following statements are equivalent:
\begin{itemize}
    \item $d$ is a Hilbertian pseudo-distance or $d$ is isometric to an $L^2-$norm.
    \item The function $ K(x, y) = e^{-\gamma d^{2\beta}(x, y)} $ is positive semidefinite for any $ \gamma \geq 0 ,  \beta \in [0, 1] $, and any $x$ and $y$ in $M$.
\end{itemize}
\end{proposition}

\subsection{Proof of Proposition~\ref{prop:sw}}
\begin{proof}
Our proof relies on the results from \cite{meunier2022slicedw}.
\begin{proposition}(Proposition 5 in \cite{meunier2022slicedw})
  The distance $\SW_2$ is Hilbertian.  
\end{proposition}
Then, applying Proposition~\ref{pos:pd_kernel}, we have  $\exp\left(-\gamma \cdot \SW_2^{2\rho}\left(\mathbb{P}(D_A), \mathbb{P}(D_B)\right)\right)$ is PSD kernel.
\begin{proposition}(Proposition~6 in \cite{meunier2022slicedw} )
  The distance $\sqrt{\SW_1}$ is a Hilbertian.  
\end{proposition}

Then, applying Proposition~\ref{pos:pd_kernel}, we have  $\exp\left(-\gamma \cdot \SW_1^{\rho}\left(\mathbb{P}(D_A), \mathbb{P}(D_B)\right)\right)$ is  a PSD kernel.

From (1), and (2) we have the proof of Proposition~\ref{prop:sw}.


\end{proof}
\subsection{Proof of Proposition~\ref{prop:ssw}}\label{app:proof_ssw}
\begin{proof}
\begin{align*}
\SSW_2^{2 \rho}(\mathbb{P}(D_A), \mathbb{P}(D_B); \mathcal{G}_{\eta}) &= {\SW}_2^{2\rho}(\mathbb{P}(\mathcal{G}_{\eta}(D_A)), \mathbb{P}(\mathcal{G}_{\eta}(D_B))) \\
&= {\SW}_2^{2\rho}(\mathbb{P}(Z_A), \mathbb{P}(Z_B)), 
\end{align*}    

where $Z_A = \mathcal{G}_{\eta}(A)$ and $Z_B = \mathcal{G}_{\eta}(B)$. Applying Proposition~\ref{prop:sw}, we have the result.

The proof for $\SW_1$ is the same.
\end{proof}

\subsection{Comparison of time complexity and properties of kernels}
\label{sec:A:compare}

\begin{table}[htbp]
\centering
\caption{Time complexity to compute dataset distances and validity for various kernels. Let $k$ denote the size of the larger dataset.}
\label{tab:dis_compare}
\begin{tabular}{c|c|c}
\toprule
\textbf{Method} & \textbf{Time Complexity} & \textbf{Valid Kernel} \\ 
\midrule
\textbf{OTDD} & \( O(k^3 \log k) \) & \xmark  \\ 
\textbf{SW}   & \( O(k \log k) \)   & \cmark \\ 
\textbf{SSW}  & \( O(k \log k) \)   & \cmark \\ 
\bottomrule
\end{tabular}
\end{table}

\section{Semivalue Estimation}
\subsection{Proof of Proposition~\ref{pos:shapley-dist}}

\begin{proof}

\textbf{Shapley Value.}
From Equation~\eqref{eq:shapley}, we rewrite
\begin{align*}
 \phi_u(i) &\triangleq \sum_{C \subseteq N \setminus i} \frac{1}{n} {\binom{n-1}{|C|}}^{-1} \left[ u(C \cup \{i\}) - u(C) \right]\\
 &= \sum_{C \subseteq N \setminus \{i\}} \left( \frac{1}{n} \binom{n - 1}{|C|} \right)^{-1} u(C \cup \{i\}) + \sum_{C \subseteq N \setminus \{i\}} \left( -\frac{1}{n} \left( \binom{n - 1}{|C|} \right)^{-1} \right) u(C)\ .\\
\end{align*}

Let $\mathcal{P}(N)$ denote the power set of $N$. Let the vector $\mathbf{u}_{\mathcal{P}(N)}$ be the values $u(C)$ for each coalition $C \subseteq N$ and $\mathbf{w}_{\mathcal{P}(N)}^{i}$ be the vector of corresponding weights for player $i$. Then, the Shapley value of player $i$ can be written compactly as
$ \phi_u(i) = \mathbf{w}_{\mathcal{P}(N)}^{i\,\top} \mathbf{u}_{\mathcal{P}(N)} $.
For each coalition $C_j \subseteq N$, the $j$-th entry of $\mathbf{w}_{\mathcal{P}(N)}^{i}$ is defined as follows:
\[
\left( \mathbf{w}_{\mathcal{P}(N)}^{i} \right)_j =
\begin{cases}
\dfrac{1}{n \binom{n-1}{|C_j|-1}} & \text{if } i \in C_j, \\
-\dfrac{1}{n \binom{n-1}{|C_j|}} & \text{if } i \notin C_j.
\end{cases}
\]

\textbf{Semivalue.}
From Equation~\eqref{eq:semi}, we rewrite
\begin{align*}
\varphi_u(i) & \triangleq \sum_{C \subseteq N \setminus i}\omega_{|C|} \left[ u(C \cup \{i\}) - u(C) \right] \\
&= \sum_{C \subseteq N \setminus \{i\}} \omega_{|C|} u(C \cup \{i\}) + \sum_{C \subseteq N \setminus \{i\}} \left(-\omega_{|C|}\right) u(C).
\end{align*}
The semivalue can also be written compactly as $\varphi_u(i) = \mathbf{w}_{\mathcal{P}(N)}^{i\,\top} \mathbf{u}_{\mathcal{P}(N)} $
where the $j$-th entry of the weight vector $\mathbf{w}_{\mathcal{P}(N)}^{i}$ is now:
\[
\left( \mathbf{w}_{\mathcal{P}(N)}^{i} \right)_j =
\begin{cases}
\omega_{|C_j|-1} & \text{if } i \in C_j, \\
-\omega_{|C_j|} & \text{if } i \notin C_j.
\end{cases}
\]

\textbf{Both.}
We can partition $\mathcal{P}(N)$ into two sets of coalitions $\cA$ and $\cB$ such that $\hat{\phi}_i$ or $\hat{\varphi}_i$ is equals to 
\[
\mathbf{w}_\cA^{i\,\top} \mathbf{u}_\cA + \mathbf{w}_{\cB}^i{}^\top \hat{\mathbf{u}}_{\cB|\cA} .
\]
Based on the GP model, the predicted utilities $\hat{\mathbf{u}}_{\cB|\cA}$ are distributed according to a multivariate Gaussian distribution. Thus, $\hat{\phi}_i$ and $\hat{\varphi}_i$ are also Gaussians.
As we use the evaluated utilities $\mathbf{u}_\cA$ of coalitions in $\cA$ and only predict the utilities $\hat{\mathbf{u}}_{\cB|\cA}$ of coalitions in $\cB$, the variance in $\hat{\phi}_i$ is only from the latter.
$\hat{\phi}_i$ and $\hat{\varphi}_i$ are Gaussians with the following mean and variance
$\mathcal{N}( \mathbf{w}_{\cA}^i{}^\top \mathbf{u}_{\cA} + \mathbf{w}_{\cB}^i{}^\top \mathbb{E}[\hat{\mathbf{u}}_{\cB|\cA}], \mathbf{w}_{\cB}^i{}^\top \mathbb{V}[\hat{\mathbf{u}}_{\cB|\cA}] \mathbf{w}_{\cB}^i ) \ .$

\end{proof}

\subsection{Discussion about Remark~\ref{rem:estimate}}\label{app:uncer_dis}
 Consider the case where our method applies Monte Carlo estimates of Shapley value to limit the number of coalitions needed for evaluation. In this scenario, the uncertainty of the Shapley value or semivalue will be bounded as $\mathbb{V}[\hat{\phi}_i']  \leq \sigma_{GP}^2 + \sigma_{MC}^2 + 2 \sigma_{GP} \sigma_{MC}$.
\begin{align*}
    \mathbb{V}[\hat{\phi}_i'] &= \mathbb{E}[(\hat{\phi}_i' - \phi_i)^2] \\
    &= \mathbb{E}[((\hat{\phi}_i' - \phi'_i) + (\phi'_i - \phi_i))^2] \\
    &= \underbrace{\mathbb{E}[(\hat{\phi}_i' - \phi'_i)^2]}_{\sigma_{GP}^2} +   \underbrace{\mathbb{E}[(\phi'_i - \phi_i)^2]}_{\sigma_{MC}^2} \\
        &+ 2\mathbb{E}[(\hat{\phi}_i' - \phi'_i)(\phi'_i - \phi_i)] \\
    & \leq \sigma_{GP}^2 + \sigma_{MC}^2 + 2 \sigma_{GP} \sigma_{MC} .
\end{align*}

\subsection{Efficiently updating the Inverse in  Algorithm~\ref{alg:active}\label{appdix:algo2}}
According to \citet{bernstein2009matrix}, Proposition 3.9.7, we have:
\begin{equation}\label{eq:inverse}
 M^{-1} = 
\begin{bmatrix}
A & B \\
C & D
\end{bmatrix}^{-1} = \begin{bmatrix}
A^{-1} + A^{-1} B S^{-1} C A^{-1} & -A^{-1} B S^{-1} \\
- S^{-1} C A^{-1} & S^{-1}
\end{bmatrix} , 
\end{equation}
where $S = D - C A^{-1} B$ is the Schur complement of $A$ in $M$.\\
In our case, when adding a new coalition $C$ to set coalition $\mathcal{A}$. We have
\[
K_{{\mathcal{A} \oplus C, \mathcal{A} \oplus C}} = \begin{bmatrix}
K_{\mathcal{A}, \mathcal{A}} & k(\mathcal{A}, C) \\
k(C, \mathcal{A}) & k(C,C) + \sigma^2
\end{bmatrix}.
\]
Then, our inverse matrix will be calculated incrementally based on Equation~\eqref{eq:inverse}.

\subsection{Complementing existing CGT-based data valuation approximations}\label{appdix:samp_algo}
How can \texttt{DUPRE} complement existing CGT-based data valuation (e.g. Shapley) approximation methods?
These Monte Carlo approximations may require the evaluation of utilities of coalitions in $\mathcal{C}$ where $|\mathcal{C}| \ll 2^n$.

We can relate $\mathcal{C}$ to the actually evaluated $\cA$ and predicted $\cB$ in Figure~\ref{fig:framework} by considering two perspectives. 
One perspective is that we can set $\mathcal{A} = \mathcal{C}$. Thereafter, we can predict the utilities of more coalitions in $\mathcal{B}$ and use them to compute another Shapley valuation approximation with more sampled coalitions.
Another perspective is that $\mathcal{C}$ is partitioned into $\mathcal{A}$ and $\mathcal{B}$. We only evaluate a subset of the coalitions and predict the remaining coalitions.

The permutation algorithm is considered in Algorithm~\ref{algo:permu_samp} and Algorithm~\ref{algo:shapley_permu_samp}. 

\begin{algorithm}[ht]
\caption{Sampling based on Permutation Sampling Algorithm}\label{algo:permu_samp}
\begin{flushleft}
\textbf{Input}: The number of coalitions to sample $n_C$, number of data owners $n_d$\\
\textbf{Output}: Selected coalitions $\mathcal{C}$, and selected permutations $\mathcal{R}$
\end{flushleft}
\begin{algorithmic}[1]
    \STATE Initialize: $\mathcal{C} \gets \emptyset$
    \STATE Initialize: $\mathcal{R} \gets \emptyset$
    \WHILE{$|\mathcal{C}| < n_C$}
        \STATE Sample a permutation $\pi$ of data owners from the uniform distribution over all permutations of $1, \cdots, n_d$
        \STATE $\mathcal{R} \gets \mathcal{R} \cup \{\pi\}$
        \STATE $S \gets \emptyset$
        \FOR{data owner $i$ in $\pi$}
            \STATE $S \gets S \cup \{\text{i}\}$
            \STATE $\mathcal{C} \gets \mathcal{C} \cup \{S\}$  
        \ENDFOR
    \ENDWHILE

\STATE return $\mathcal{C}, \mathcal{R}$
\end{algorithmic}
\end{algorithm}

\begin{algorithm}[ht]
\caption{Shapley Estimate based on Permutation Sampling Algorithm}\label{algo:shapley_permu_samp}
\begin{flushleft}
\textbf{Input}: List of all permutations $\mathcal{R}$, list of actually evaluated coalitions $\cA$, list of predicted coalitions $\cB$, the utility function $u$\\
\textbf{Output}: List of Shapley values $(\phi_u(i))_{i \in N}$
\end{flushleft}
\begin{algorithmic}[1]
    \STATE Actually evaluate the utilities $\mathbf{u}_\cA \triangleq (u(A))_{A \in \cA}$ for each coalition $A$ in $\cA$
    \STATE Predict the utilities $ \mathbb{E}[\mathbf{\hat{u}}_{\cB|\cA}] =
  \mathbf{K}_{\cB,\cA} [\mathbf{K}_{\cA,\cA} + \sigma^2 \mathbf{I}]^{-1} \mathbf{u}_\cA $ for coalitions in $\cB$
    \FOR{$\pi$ in $\mathcal{R}$}
        \STATE $prev \gets 0$
        \STATE $S \gets \emptyset$ 
        \FOR{$i$ in $\pi$}
            \STATE $S \gets S \cup \{i\}$ 
            \IF{$S \in \cA$}
            \STATE $curr \gets u(S)$ saved in $\mathbf{u}_\cA$ 
            \ELSIF{$S \in \cB$}
            \STATE $curr \gets \hat{u}(S)$ saved in $ \mathbb{E}[\mathbf{\hat{u}}_{\cB|\cA}]$
            \ENDIF
            \STATE $\phi_{u}(i) \gets \phi_{u}(i) + (curr - prev) / |\mathcal{R}|$
            \STATE $prev \gets curr$
        \ENDFOR
    \ENDFOR
\STATE return $(\phi_u(i))_{i \in N}$
\end{algorithmic}
\end{algorithm}

\clearpage

\section{Experiments}
\label{appendix:exp}

\subsection{Detailed Experiment Setup}

\textbf{Datasets and Models} Table~\ref{tab:datasize_per_owner} summarizes the details of the datasets, models, and hardware used in each experiment. Specifically, NN refers to a neural network architecture with three hidden layers, and MLP stands for a Multi-Layer Perceptron regressor with three hidden layers and two ReLU activation layers. For the MNIST custom data division, based on different numbers of data owners, the division will be \{0,1\}, \{2,3,4\}, \{3,4,5\}, \{6,7\}, \{8,9\} and $\{1, 4, 5, 7, 8\}, \{2\}, \{9\}, \{6\}, \{0\}$ 

\textbf{Software}: We use Python libraries, including, PyTorch and \texttt{pyDVL} \cite{TransferLab_team_pyDVL_2024}.

\textbf{Hardware}: We primarily run experiments on NVIDIA GeForce RTX 3080 (10GB) and NVIDIA L40 (40GB) GPUs.

\textbf{Training Procedure}: The ML model was trained for 100 epochs with a learning rate of 0.001. For CaliH and Moon Dataset, we use full-batch training. In contrast, for the CIFAR-10 and MNIST datasets, the batch sizes were set to 256 and 64, respectively. We run every experiment ten times with seed from 0 to 9.
\begin{table}[h]
\centering
\caption{Overview of Experiments and Datasets}
\label{tab:datasize_per_owner}
\begin{tabular}{|l|c|c|c|c|c|l|}
\hline
\textbf{Dataset} & \textbf{Num data-owners} & \textbf{Data Size per Owner}&  \textbf{Train/Valid} &  \textbf{Division} &  \textbf{ML model} & \textbf{Exp} \\ \hline
MNIST            & 10                             & 6,000                    & 60k/10k& per digit&  NN&   Section~\ref{sec:est_sampling_shapley} \\ \hline
CaliH            & 6                              & 2,700                    & 16k/4k  & random &  MLP& Section~\ref{sec:cup}\\ \hline
MNIST            & 6                             & 20,000 or 6,000            & 60k/10k&  custom &   NN&
Section~\ref{sec:cup}\ \\ \hline
MNIST            & 5                              & 12,000 (lowest) or 18,000    & 60k/10k  & custom &  NN& Section~\ref{sec:benfun} \& Section~\ref{subsec:ablation} \\ \hline
CIFAR            & 8                              & 5,000 or 10,000      & 50k/10k  & per label& Resnet-18 &Section~\ref{sec:exp_exact_shap}\\ \hline
IMDb            & 10                              & 2,500    & 25k/25k  & random& Resnet-18 &Section~\ref{subsec:nlp-datasets}\\ \hline
IMDb            & 20                              & 100 to 2000    & 25k/29k  & custom& Resnet-18 &Section~\ref{subsec:imdb-robustness}\\ \hline
\end{tabular}
\end{table}

\subsection{Experiments on an Additional NLP Dataset (SST-2)}
\label{appendix:SST-2}
In this section, we conducted additional experiments on the Stanford Sentiment Treebank (SST-2) involves classifying the sentiment of movie reviews as either positive or negative. We use a similar setting to Section~\ref{subsec:nlp-datasets} on 10 data owners. Following the OpenDataVal benchmark, we use DistilBERT embeddings for each review. We consider two evaluation setups:
\begin{itemize}
  \item \textbf{Setup~1 (Section \ref{sec:cup})} uses 256 evaluated coalitions and is used to study the agreement between the actual utility $u(C)$ and GP predicted utility $\hat{u}(C)$ of various coalitions $C$.
  \item \textbf{Setup~2 (Section \ref{sec:est_sampling_shapley})} uses 512 evaluated and 100 predicted coalitions and is used to study the agreement between the actual Shapley value $(\phi_i)_{i \in N}$ and the predicted Shapley value $(\hat{\phi}_i)_{i \in N}$. 
\end{itemize}

\begin{table*}[h!]
\centering
\caption{
\textbf{A comparison of the quality of utility predictions and Shapley value predictions for various methods on the SST-2 dataset with $10$ owners.}
\textbf{Setup~1 (similar to Section \ref{sec:cup})} compares the actual utility $u(C)$ with GP predicted utility $\hat{u}(C)$ of various coalition $C$.
\textbf{Setup~2 (similar to Section \ref{sec:est_sampling_shapley})} compares the actual Shapley value $\phi_i$ and the predicted Shapley value $(\hat{\phi}_i)$ of various owner $i$. 
The results are the mean $\pm$ std.\ over 5 runs.
Lower MSE and higher correlation are preferred.
}
\label{tab:sst2-results}
\begin{tabular}{l|ccc|cc}
\toprule
& \multicolumn{3}{c|}{\textbf{Setup 1 (Section \ref{sec:cup})}} 
& \multicolumn{2}{c}{\textbf{Setup 2 (Section \ref{sec:est_sampling_shapley}) }} \\
\textbf{Method} 
& $\text{MSE}(u(C), \hat{u}(C))$ 
& $\text{Pearson}\bigl(u(C), \hat{u}(C) \bigr)$
& $\text{Shapley Corr.}$
& $\text{MSE}((\phi_i)_{i \in N}, (\hat{\phi}_i)_{i \in N})$
& $\text{Pearson}((\phi_i)_{i \in N}, (\hat{\phi}_i)_{i \in N})$ \\
\midrule
\texttt{SSW (Ours)} 
& \(\mathbf{2.024\times10^{-5}\pm0.00012}\)
& \(\mathbf{0.93\pm0.04}\)
& \(\mathbf{0.91\pm0.06}\)
& \(\mathbf{0.0002\pm0.0038}\)
& \(\mathbf{0.50\pm0.234}\)\\

\texttt{GP-binary}  
& \(2.638\times10^{-5}\pm0.00018\)
& \(0.733\pm0.05\)
& \(0.84\pm0.06\)
& \(0.00033\pm0.0042\)
& \(0.35\pm0.26\) \\

\texttt{NN-binary}  
& \(0.005\pm0.007\) 
& \(0.422\pm0.19\)
& \(0.334\pm0.29\)
& \(0.00025\pm0.007\)
& \(0.37\pm0.186\) \\
\bottomrule
\end{tabular}
\end{table*}

In Table~\ref{tab:sst2-results}, \texttt{SSW} still leads to the lowest MSE and the highest correlation as compared to other kernels.

\subsection{Experiments on Another Semivalue: Banzhaf value}
\label{appendix:banzhaf}
We reuse the setup of Section~\ref{sec:exp_exact_shap} but consider computing the Banzhaf value instead of the Shapley value. In Table~\ref{tab:compare_corr_banzhaf}, we observe that the kernel based on $\texttt{SSW}$ leads to the highest correlation between the actual and predicted Banzhaf values.
\begin{table*}[ht]
\caption{A comparison of the quality of exact Banzhaf value approximation on the CIFAR-10 and CaliH datasets with $8$ data owners. A higher correlation is preferred.}
\centering
\label{tab:compare_corr_banzhaf}
\begin{tabular}{l|cc|cc}
    \toprule
    \multirow{2}{*}{\textbf{Method}} 
    & \multicolumn{2}{c}{\textbf{CIFAR-10}} 
    & \multicolumn{2}{c}{\textbf{CaliH}} \\ 
    & \textbf{Pearson} & \textbf{Kendall Tau} 
    & \textbf{Pearson} & \textbf{Kendall Tau}\\ 
    \midrule
    \texttt{SSW} &  $\mathbf{0.98 \pm 0.016} $    & $\mathbf{0.70 \pm 0.15}$ &     $\mathbf{0.97 \pm 0.02}$ & $\mathbf{0.9 \pm 0.069}$\\ 
    \texttt{OTDD} (invalid kernel baseline) &  $0.69 \pm 0.082$&  $0.52 \pm 0.14$ &   - &-    \\ 
    \texttt{GP-binary} (baseline 01 encoding) & $0.92 \pm 0.03$ &  $0.66 \pm 0.18$ & $0.95 \pm 0.06$ & $0.85 \pm 0.11$  \\ 
    \texttt{NN-binary} (baseline 01 encoding) &  $0.65 \pm 0.01$   & $0.57 \pm 0.14$  & $0.87 \pm 0.09$ & $0.72 \pm 0.062$ \\ 
    \texttt{LAVA} (baseline) &  $0.0034 \pm 0.045$& $-0.078 \pm 0.1$ & $0.135 \pm 0.147$ &  $0.028 \pm 0.15$   \\ 
    \bottomrule
\end{tabular}
\end{table*}
\fi
\end{document}